\documentclass[letterpaper, 10 pt, conference]{ieeeconf}  

\IEEEoverridecommandlockouts                              

\usepackage{cite}
\usepackage{amsmath,amsfonts}
\usepackage{algorithmic}
\usepackage{algorithm}
\usepackage{stfloats}
\usepackage{array}
\usepackage{booktabs}
\usepackage{verbatim}
\usepackage{graphicx}  
\usepackage[caption=false,font=footnotesize,labelfont=rm,textfont=rm]{subfig}
\newcommand{\upcite}[1]{\textsuperscript{\textsuperscript{\cite{#1}}}}
\title{\LARGE \bf
From Learning to Mastery: Achieving Safe and Efficient Real-World Autonomous Driving with Human-in-the-Loop Reinforcement Learning
}
\author{Zeqiao Li,~Yijing Wang, ~Haoyu Wang, ~Zheng Li, ~Peng Li, ~Wenfei Liu, ~Zhiqiang Zuo
\thanks{*This work was supported in part by the National Natural Science Foundation of China under Grant 62173243, Grant 62403348, and the Young Scientists Fund of the National Natural Science Foundation of Tianjin, China, under Grant 23JCQNJC01780, and the Postdoctoral Fellowship Program of CPSF under Grant GZC20241208, Grant 2024M762357, and the Foundation of Key Laboratory of System Control and Information Processing, Ministry of Education, P.R. China, under Grant Scip20240116.}
\thanks{The authors are with the Tianjin Key Laboratory of Intelligent Unmanned Swarm Technology and System, School of Electrical and Information Engineering, Tianjin University, Tianjin 300072, China;~Haoyu Wang is also with Key Laboratory of System Control and Information Processing, Ministry of Education of China, Shanghai, 200240 ({\tt\small email: lizeqiao@tju.edu.cn; yjwang@tju.edu.cn; why2014@tju.edu.cn; zhengl@tju.edu.cn; lipeng\_2017@tju.edu.cn; liuwenfei@tju.edu.cn; zqzuo@tju.edu.cn}).
}
}

\begin{document}

\maketitle
\thispagestyle{empty}
\pagestyle{empty}

\begin{abstract}
  Autonomous driving with reinforcement learning (RL) has significant potential. However, applying RL in real-world settings remains challenging due to the need for safe, efficient, and robust learning. Incorporating human expertise into the learning process can help overcome these challenges by reducing risky exploration and improving sample efficiency. In this work, we propose a reward-free, active human-in-the-loop learning method called Human-Guided Distributional Soft Actor-Critic (H-DSAC). Our method combines Proxy Value Propagation (PVP) and Distributional Soft Actor-Critic (DSAC) to enable efficient and safe training in real-world environments. The key innovation is the construction of a distributed proxy value function within the DSAC framework. This function encodes human intent by assigning higher expected returns to expert demonstrations and penalizing actions that require human intervention. By extrapolating these labels to unlabeled states, the policy is effectively guided toward expert-like behavior. With a well-designed state space, our method achieves real-world driving policy learning within practical training times. Results from both simulation and real-world experiments demonstrate that our framework enables safe, robust, and sample-efficient learning for autonomous driving. The videos and code are available at: https://github.com/lzqw/H-DSAC.

\end{abstract}

\section{INTRODUCTION}
Autonomous driving (AD) has the potential to revolutionize transportation by enhancing road safety, alleviating traffic congestion, and expanding mobility\cite{SurveyRLIL02}. However, developing a robust AD system is highly challenging due to the need to navigate dynamic and uncertain environments, address perception errors that impact decision-making, and ensure safe, efficient real-time decisions within high-dimensional state spaces\cite{RLSurvey03}. Reinforcement learning (RL) offers a promising approach, enabling agents to autonomously acquire driving skills through direct environmental interaction\cite{IROS2}. By incorporating objectives such as safety and efficiency into reward functions, RL provides a flexible framework for learning a wide range of driving tasks, from basic lane-keeping to complex urban maneuvers. Despite its potential, traditional RL methods face several limitations, including poor sample efficiency and risky trial-and-error exploration, which hinder their practical deployment in real-world applications. Addressing these challenges requires advanced techniques to mitigate unsafe interactions and accelerate policy learning. Enhancing sample efficiency and ensuring safer training in RL-based approaches are crucial steps toward making reinforcement learning a viable and scalable solution for autonomous driving.\\
\indent RL in autonomous driving faces several challenges. Poor sample efficiency often necessitates extensive data collection, which is costly and risky, especially for rare but critical events such as sudden lane changes or emergency braking. This restricts RL's ability to efficiently learn important but infrequent behaviors. Safety during training is another major concern, as trial-and-error exploration can lead to unsafe maneuvers, highlighting the need for safeguards to reduce collisions or near-misses\cite{RLSurvey01}. Reward design is also highly complex, as driving tasks involve balancing multiple objectives like safety, comfort, and efficiency. Poorly designed reward functions may lead to unintended, unsafe actions. Additionally, the sim-to-real transfer poses a significant hurdle\cite{RLReward01}. Models trained in simulation often suffer from performance degradation when deployed in the real world due to differences in lighting, textures, dynamics, and sensor noise\cite{IROS3}. Addressing these challenges requires developing algorithms that improve sample efficiency, enhance safety, and ensure effective deployment across both simulated and real environments.\\
\indent Human experts possess deep insights into the tasks performed by agents, which significantly enhances exploration efficiency and reduces reliance on trial-and-error learning\cite{HILSHORT02}. To tackle the pervasive issue of low sample efficiency in RL, various human-in-the-loop RL (HIL) methods have been proposed. The core principle of HIL is to establish a feedback loop between the learning agent and human experts\cite{IROS4}. For example, human experts actively participate in the training process, iteratively refining the learned policy\cite{DAGGER}. Some approaches allow the agent to request human guidance when needed\cite{IROS5}, while others involve human experts providing preference-based feedback on collected trajectories\cite{IROS7}. These methods not only improve sampling efficiency but also mitigate the challenge of designing complex reward functions. However, in high-stakes domains such as autonomous driving, ensuring safety during training remains a critical challenge. To address this, certain methods enable human experts to actively intervene and provide demonstrations during execution\cite{EIL,HACO}. While these strategies can accelerate learning and enhance policy interpretability, they also introduce new challenges. A crucial concern is the burden placed on human experts due to their involvement in the training process. Therefore, it is essential to develop methods that effectively capture and represent human guidance while minimizing cognitive load, ensuring both safety and efficiency in learning human intentions.\\
\indent In this paper, we propose a Human-Guided Distributional Soft Actor-Critic (H-DSAC) method for real-world autonomous driving. By integrating Proxy Value Propagation into the Distributional Soft Actor-Critic (DSAC) algorithm, our scheme combines human guidance with DSAC to improve sample efficiency and enhance safety during training. A key feature of our approach is the distributional proxy value function, which captures human intent through return distributions and guide policy learning to mimic human behaviors. These distributed proxy values are propagated to unlabeled state-action pairs during the agent’s exploration, leveraging temporal-difference (TD) learning within DSAC. This strategy enables the agent to acquire fundamental driving skills both efficiently and safely. Our method strikes a balance between human expertise and autonomous discovery, resulting in faster and safer learning.\\
\indent Our contributions can be summarized as follows:
\begin{itemize}
\item We put forward a distributional proxy value function that encodes human intent through return distributions. This function guides policy learning by assigning higher returns to expert-like actions and lower returns to those that require human intervention, thereby ensuring safer and more efficient learning.
\item We propose the H-DSAC, which integrates human feedback with off-policy RL. This approach enhances sample efficiency, accelerates policy convergence, and improves safety during training, enabling effective learning from both human demonstrations and autonomous exploration.
\item Our framework allows the vehicle to learn driving strategies directly in real environments within practical training times. By leveraging robust state representations and incorporating H-DSAC, it ensures an efficient and safe learning process, enabling real-time training in real-world conditions.
\end{itemize}

\section{RELATED WORK}
This section reviews the existing research across key areas relevant to our work: reinforcement learning (RL) and human-in-the-loop reinforcement learning (HIL).
\subsection{Reinforcement Learning}
RL, as a powerful paradigm for training autonomous systems through trial-and-error interactions, enables agents to establish causal relationships among observations, actions, and outcomes \cite{RLSurvey01}. \cite{IROS8} proposed the first RL algorithm suitable for continuous control settings, known as Deep Deterministic Policy Gradient (DDPG), and successfully implemented a lane-keeping function using simulated images as input on the TORCS driving simulation platform \cite{IROS8}. Since then, a number of mainstream RL algorithms, including DDPG \cite{IROS9}, Asynchronous Advantage Actor-Critic (A3C) \cite{IROS10}, and Proximal Policy Optimization (PPO) \cite{IROS11}, have been employed to achieve similar driving functions. The majority of these studies have been carried out in simulation environments such as TORCS and CARLA. However, verifying the effectiveness of the learned policy on a real vehicle is of paramount importance. To address the challenges in RL-based driving plicy learning, Duan et al. introduced Distributional Soft Actor-Critic (DSAC) and its variant DSAC-T, which mitigate Q-value overestimation by modeling the distribution of state-action returns, thereby enhancing policy performance \cite{DSAC, DSAC-T}. In \cite{DSAC-REAL}, DSAC was applied to the highway on-ramp merging decision-making problem, integrating a safety shield based on barrier functions for online corrections. This approach not only enhances merging efficiency but also ensures safety. Despite these advancements, training in simulation before deployment still faces challenges related to the sim-to-real gap and adaptability. As a result, some studies have shifted focus to real-world RL. Several algorithms have demonstrated the capability to learn efficiently in real-world scenarios \cite{REAL1, REAL2, REAL3}. However, real-world RL methods often require extensive training time, which poses practical limitations on their deployment.\\
\subsection{Human-in-the-Loop Reinforcement Learning}
HIL strategies aim to mitigate the risk of unsafe exploration by integrating human expertise directly into the learning loop. \cite{DAGGER} proposed an iterative algorithm called DAgger, which trains a stationary deterministic policy. This can be viewed as a no regret algorithm in an online learning setting. As the extensions of DAgger, \cite{DAGGEROther01,DAGGEROther02,DAGGEROther03} enable the human expert to intercede during exploration and guide the agent back to secure states thereby mitigating the compounding effect of incorrect actions. Expert Intervention Learning (EIL) \cite{EIL} and Intervention Weighted Regression (IWR) \cite{IWR} allow human operators to take over control during high-risk situations, steering the agent toward safer states. Other methods collect human evaluative feedback on agent-generated trajectories to ensure alignment with human preferences \cite{HIL01,HIL02,HIL03}. Recent advances like HACO \cite{HACO} dynamically adjust autonomy levels to reduce the burden of continuous human supervision. This is achieved through reliance on partial demonstrations and limited interventions for data collection. Meanwhile, Proxy Value Propagation (PVP) \cite{PVP} encodes human intentions into a proxy value function, efficiently guiding agents toward behavior patterns that align human judgment. Despite these innovations, HIL approaches still face significant challenges. Continuous human oversight also places heavy demands on operators, complicating large-scale deployment \cite{HILSHORT02}. Balancing human guidance, safety, and efficient policy learning thus remains a critical challenge in advancing HIL-based driving systems.\\
\begin{figure*}[!t]
  \centering
  \includegraphics[width=7.in]{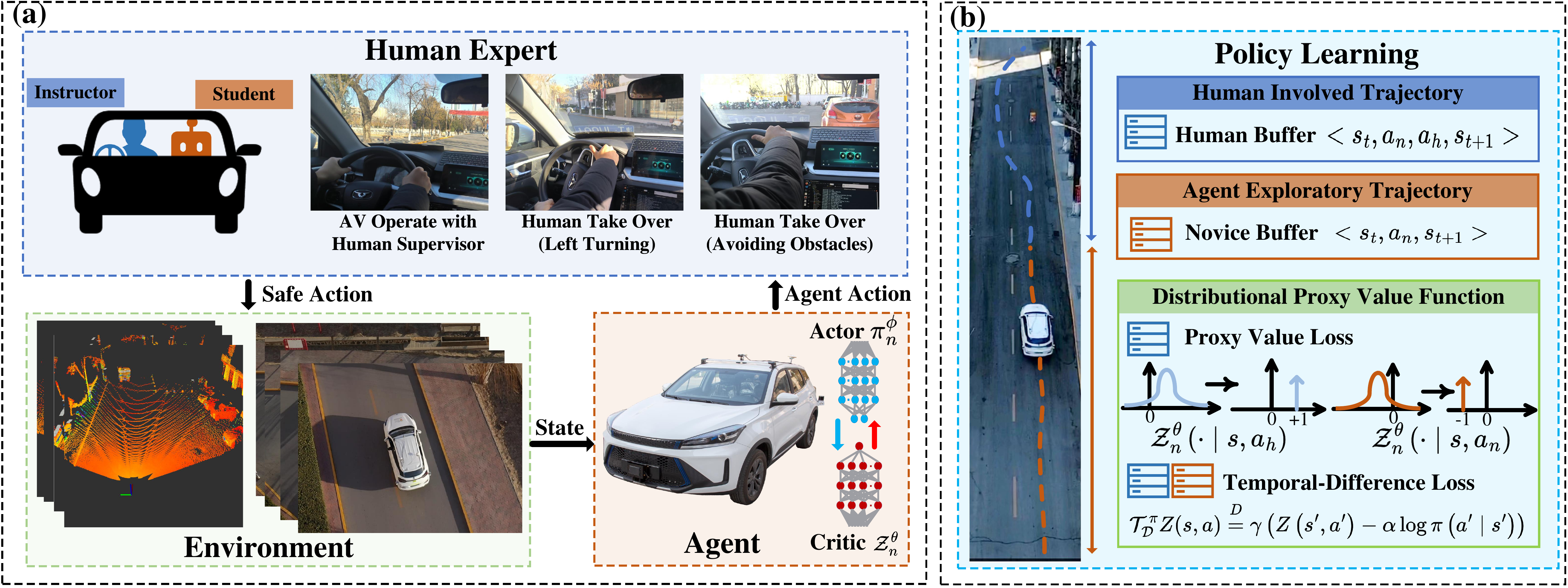}
  \caption{Overall framework of H-DSAC}
  \label{fig_framework}
  \end{figure*}
\section{Methodology}
In this section, we present the problem formulation of end-to-end autonomous driving and provide a detailed introduction to H-DSAC. Then, we elaborate the simulation experiment setup on the MetaDrive safety benchmark and the real-world experiment design on an Unmanned Ground Vehicle (UGV) platform. 
\subsection{Problem Statement}
The policy learning of end-to-end autonomous driving can be framed as a continuous action space problem within the realm of RL. Specifically, this problem can be formulated as a Markov Decision Process (MDP) $\mathcal{M}=\left(\mathcal{S}, \mathcal{A}, \mathcal{P}, \mathcal{R}, \gamma\right)$, where $\mathcal{S}$ denotes the state space, $\mathcal{A}$ is the action space, $\mathcal{P}$ represents the transition probability, $\mathcal{R}$ is the reward function, and $\gamma$ stands for the discount factor. The objective in standard RL is to learn a policy $\pi: \mathcal{S} \rightarrow \mathcal{A}$ that maximizes the expected cumulative reward $R_t = \sum_{t=0}^{\infty} \gamma^t r_t$, with $r_t$ being the reward at time $t$. In this study, we consider an entropy-augmented objective function\cite{SAC}, which incorporates policy entropy into the reward term:
\begin{equation}
  J_\pi=\underset{\left(s_{i \geq t}, a_{i \geq t}\right) \sim \rho_\pi}{\mathbb{E}}\left[\sum_{i=t}^{\infty} \gamma^{i-t}\left[r_i+\alpha \mathcal{H}\left(\pi\left(\cdot \mid s_i\right)\right)\right]\right],
  \label{entropy_augmented_objective_function}
  \end{equation}
where $\alpha$ is the temperature coefficient, and the policy entropy $\mathcal{H}$ is expressed as
\begin{equation}
  \mathcal{H}(\pi(\cdot \mid s))=\underset{a \sim \pi(\cdot \mid s)}{\mathbb{E}}[-\log \pi(a \mid s)].
  \end{equation}
\indent The soft $Q$ value is given by
\begin{equation}
  \begin{aligned}
  Q^\pi\left(s_t, a_t\right)&=r_t\\
  +\gamma &\underset{\substack{\left(s_{i>t}, a_{i>t}\right) \sim \rho_\pi}}{\mathbb{E}}\left[\sum_{i=t}^{\infty} \gamma^{i-t}\left[r_i-\alpha \log \pi\left(a_i \mid s_i\right)\right]\right],
\end{aligned}
\label{soft_q_value}
\end{equation}
which delineates the expected soft return for choosing $a_t$ at state $s_t$ under policy $\pi$. \\
\indent In HIL, when the agent encounters a risky situation or makes a suboptimal decision, the human expert can execute an action $a_h$ to overwrite the agent's action $a_n$. This intervention mechanism allows the action $a_b$ applied to the environment to be written as:
\begin{equation}
a_b = I(s, a) a_h + (1 - I(s, a)) a_n
\end{equation}
where $I(s, a)$ is a boolean indicator function.\\
\indent With the human policy $\pi_h$ and the agent's policy $\pi_n$, the policy $\pi_b$ used for generating the actual trajectory is defined as:
\begin{equation}
\pi_b(a \mid s) = \pi_n(a \mid s) (1 - I(s, a)) + \pi_h(a \mid s) G(s)
\end{equation}
where $G(s)$ is the probability of human intervention and it has the form:

\begin{equation}
G(s) = \int_{a' \in \mathcal{A}} I(s, a') \pi_h(a' \mid s) da'
\end{equation}

\subsection{Human-Guided Distributional Soft Actor-Critic}
As shown in Fig. \ref{fig_framework}(a), the H-DSAC adheres to the HIL setup. The agent interacts with the environment and collects data, which is stored in the novice buffer $\mathcal{B}_n$. The human supervisor can intervene at any time, providing expert demonstrations that are recorded in the human buffer $\mathcal{B}_h$. During the initial training phase, the novice policy $\pi_n$ is initialized randomly, while the human policy $\pi_h$ is treated as a fixed one. Early in training, human intervention is dominant, and the novice policy is updated using the H-DSAC, which integrates human demonstrations to guide policy learning. As training progresses, the novice policy gradually converges towards the expert policy, reducing the frequency of human intervention decreases. Ultimately, this enables the agent to achieve autonomous driving capability. \\
\indent The H-DSAC is designed to efficiently guide the learning of the novice policy by leveraging expert demonstrations. The core idea behind H-DSAC is to propagate human feedback through a distributional proxy value function. This function captures not only the expected return but also the variability of outcomes, thereby enabling the agent to better handle uncertainty in its learning process. \\
\indent We first define the soft state-action return as:
\begin{equation}
  Z^\pi(s_t, a_t) := r_t + \gamma \sum_{i=t}^{\infty} \gamma^{i-t} \left[r_i - \alpha \log \pi(a_i \mid s_i)\right].
\end{equation} 
\indent Then let the distribution of the random variable $Z^\pi(s, a)$ be $\mathcal{Z}^\pi(Z^\pi(s, a) \mid s, a)$. Accordingly, the reward-free and distributional version of the soft bellman operator is formulated as:
\begin{equation}
  \mathcal{T}_{\mathcal{D}}^\pi Z(s, a) \stackrel{D}{=} \gamma \left( Z(s', a') - \alpha \log \pi(a' \mid s') \right),
  \label{reward_free_distributional_version_of_the_soft_Bellman_operator}
\end{equation}
where $s' \sim p$, $a' \sim \pi$, and $A \stackrel{D}{=} B$ signifies that the two random variables $A$ and $B$ share identical probability distributions.\\
\indent In H-DSAC, there is a distributional value network and a stochastic policy, parameterized by $\mathcal{Z}^{\theta}_n(\cdot \mid s, a)$ and $\pi^{\phi}_n(\cdot \mid s)$, respectively. The distribution $\mathcal{Z}_n^{\theta}$ is specifically designed for proxy value propagation and reward-free TD learning. Both networks are modeled as diagonal Gaussian distributions, outputting the mean and standard deviation.\\
\indent For distributional proxy value function, as illustrated in Fig. \ref{fig_framework}(b), the objective is to emulate human behavior while minimizing the need for  intervention. It samples data $\left(s, a_n, a_h\right)$ from the human buffer and assigns value distributions to the human action $a_h$ and the novice action $a_n$. The value distribution of the human action $a_h$ is labeled as $\delta_1(\cdot)$, while the novice action $a_g$ is labeled with $\delta_{-1}(\cdot)$. Here, $\delta_1(\cdot)$ and  $\delta_{-1}(\cdot)$ represent Dirac delta distributions centered at 1 and -1, respectively. This labeling scheme is designed to fit $\mathcal{Z}_n^{\theta}(\cdot \mid s, a)$ through the following distributional proxy value (PV) loss:
\begin{equation}
  J_{\mathcal{Z}}^{PV}(\theta) = \left( J_{\mathcal{Z}}^H(\theta) + J_{\mathcal{Z}}^N(\theta) \right) I(s, a_n),
\end{equation}
where
\begin{equation}
  J_{\mathcal{Z}}^H(\theta) = \mathbb{E}_{(s, a_h, a_n) \sim \mathcal{B}_h}\left[D_{\mathrm{KL}}\left(\delta_1(\cdot), \mathcal{Z}_n^{\theta}(\cdot \mid s, a_h)\right)\right]
\end{equation}
and
\begin{equation}
  J_{\mathcal{Z}}^N(\theta) = \mathbb{E}_{(s, a_h, a_n) \sim \mathcal{B}_h}\left[D_{\mathrm{KL}}\left(\delta_{-1}(\cdot), \mathcal{Z}_n^{\theta}(\cdot \mid s, a_n)\right)\right].
\end{equation}
\indent Since $\mathcal{Z}_n^{\theta}$ is Gaussian, $\mathcal{Z}_n^{\theta}(\cdot\mid s, a)$ can be expressed as $\mathcal{N}(Q_\theta(s, a), \sigma_\theta(s, a)^2)$, where $Q_\theta(s, a)$ and $\sigma_\theta(s, a)$ are the mean and standard deviation of the return distribution. The update gradient for $J_{\mathcal{Z}}^H(\theta)$ and $J_{\mathcal{Z}}^N(\theta)$ are:

\begin{equation}
  \begin{aligned}
  \nabla_\theta J_{\mathcal{Z}}^H(\theta)=&\mathbb{E}\left[\nabla_\theta \frac{\left(1-Q_\theta(s, a)\right)^2}{2 \sigma_\theta(s, a)^2}+\eta \frac{\nabla_\theta \sigma_\theta(s, a)}{\sigma_\theta(s, a)}\right] \\
  =&\mathbb{E}\left[-\frac{\left(1-Q_\theta(s, a)\right)}{\sigma_\theta(s, a)^2} \nabla_\theta Q_\theta(s, a) \right.\\
  &\left.-\frac{\left(1-Q_\theta(s, a)\right)^2-\sigma_\theta(s, a)^2}{\sigma_\theta(s, a)^3} \eta \nabla_\theta \sigma_\theta(s, a)\right],
  \end{aligned}
  \end{equation}
  
    and
    \begin{equation}
      \begin{aligned}
      \nabla_\theta J_{\mathcal{Z}}^N(\theta)=&\mathbb{E}\left[\nabla_\theta \frac{\left(1-Q_\theta(s, a)\right)^2}{2 \sigma_\theta(s, a)^2}+\eta \frac{\nabla_\theta \sigma_\theta(s, a)}{\sigma_\theta(s, a)}\right] \\
      =&\mathbb{E}\left[\frac{\left(1-Q_\theta(s, a)\right)}{\sigma_\theta(s, a)^2} \nabla_\theta Q_\theta(s, a)\right.\\
  &\left.-\frac{\left(1-Q_\theta(s, a)\right)^2-\sigma_\theta(s, a)^2}{\sigma_\theta(s, a)^3} \eta \nabla_\theta \sigma_\theta(s, a)\right].
      \end{aligned}
      \end{equation}
where $\eta$ modulates the variance convergence rate.\\
\indent Transitions stored in the novice buffer, though devoid of human intervention, still encapsulate valuable information regarding forward dynamics and human preferences. Instead of discarding these data, H-DSAC propagates proxy values to these states through a reward-free TD update. As illustrated in Fig. \ref{fig_framework}(b), the reward-free TD loss is defined as:

\begin{equation}
  J^{TD}_{\mathcal{Z}}(\theta) = \mathbb{E}_{(s, a) \sim \mathcal{B}}\left[D_{\mathrm{KL}}\left(\mathcal{T}_{\mathcal{D}}^{\pi_n^{\bar{\phi}}} \mathcal{Z}_n^{\bar{\theta}}(\cdot \mid s, a), \mathcal{Z}_n^\theta(\cdot \mid s, a)\right)\right],
  \label{reward_free_td_loss}
\end{equation}
where $\bar{\theta}$ and $\bar{\phi}$ denote the target network parameters, and $\mathcal{B}$ represents the union of the novice and human buffers, $\mathcal{B}_n \cup  \mathcal{B}_h$. Since the term $\mathcal{T}_{\mathcal{D}}^{\pi_n^{\bar{\phi}}} \mathcal{Z}_n^{\bar{\theta}}$ is not explicitly available, we approximate the computation using a sample-based formulation:
\begin{equation}
  J^{TD}_{\mathcal{Z}}(\theta)=-\underset{\substack{(s, a) \sim \mathcal{B}\\Z\left(s^{\prime}, a^{\prime}\right) \sim \mathcal{Z}_n^{\bar{\theta}}\left(\cdot \mid s^{\prime}, a^{\prime}\right)}}{\mathbb{E}}\left[\log \mathcal{P}\left(y_z \mid \mathcal{Z}_n^\theta(\cdot \mid s, a)\right)\right],
  \label{kl_update}
  \end{equation}
  where the reward-free target value is given by:
\begin{equation}
  y_z=\gamma\left(Z\left(s^{\prime}, a^{\prime}\right)-\alpha \log \pi^g_{\bar{\phi}}\left(a^{\prime} \mid s^{\prime}\right)\right).
  \end{equation}

\indent And the corresponding gradient update is expressed as:

\begin{equation}
  \begin{aligned}
  \nabla_\theta J_{\mathcal{Z}}^{TD}(\theta)=&\mathbb{E}\left[\nabla_\theta \frac{\left(y_z-Q_\theta(s, a)\right)^2}{2 \sigma_\theta(s, a)^2}+\eta \frac{\nabla_\theta \sigma_\theta(s, a)}{\sigma_\theta(s, a)}\right] \\
  =&\mathbb{E}\left[-\frac{\left(y_z-Q_\theta(s, a)\right)}{\sigma_\theta(s, a)^2} \nabla_\theta Q_\theta(s, a)\right. \\
  &\left.-\frac{\left(y_z-Q_\theta(s, a)\right)^2-\sigma_\theta(s, a)^2}{\sigma_\theta(s, a)^3} \eta \nabla_\theta \sigma_\theta(s, a)\right]
  \end{aligned}
  \end{equation}

\indent The final value loss for $\mathcal{Z}_n^\theta(\cdot \mid s, a)$ integrates both distributional proxy value loss (PV) loss and TD loss, ensuring effective value propagation:

\begin{equation}
  J_{\mathcal{Z}}(\theta) = J_{\mathcal{Z}}^{PV}(\theta) + J_{\mathcal{Z}}^{TD}(\theta)
  \label{final_value_loss}
\end{equation}

\indent For policy improvement, the actor $\pi_n^\phi(\cdot \mid s)$ is updated by maximizing the return distribution:

\begin{equation}
\begin{aligned}
  J_{\pi}(\phi) & = \underset{\substack{s \sim \mathcal{B}\\ a \sim \pi_n^\phi}}{\mathbb{E}}\left[\underset{\substack{Z(s, a) \sim \mathcal{Z}_n^\theta(\cdot \mid s, a)}}{\mathbb{E}}[Z(s, a)] -\alpha \log \left(\pi_n^\phi(a \mid s)\right)\right] \\
  & = \underset{s \sim \mathcal{B}, a \sim \pi_n^\phi}{\mathbb{E}}\left[Q_\theta(s, a) - \alpha \log \left(\pi_n^\phi(a \mid s)\right)\right],
\end{aligned}
\label{policy_improvement}
\end{equation}
\indent To maintain an appropriate balance between exploration and exploitation, the temperature parameter $\alpha$ is adaptively adjusted:

\begin{equation}
  \alpha \leftarrow \alpha - \mathbb{E}_{s \sim \mathcal{B}, a \sim \pi_n^\phi}\left[-\log \pi_n^\phi(a \mid s) - \overline{\mathcal{H}}\right]
\end{equation}

\subsection{Simulation Experiment Design}

\begin{figure}[]
  \centering
  \includegraphics[width=3.2in]{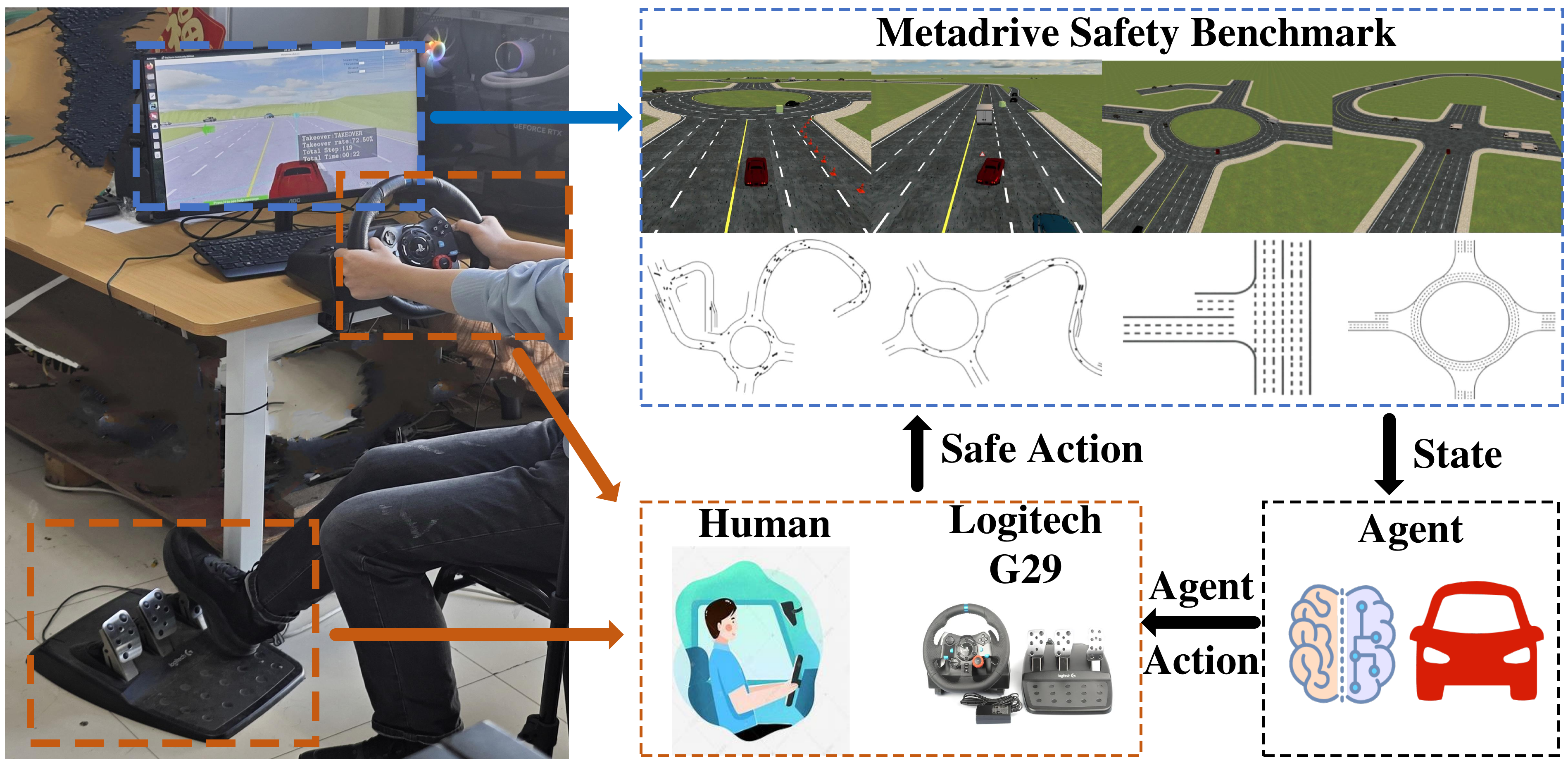}
  \caption{Simulation environment and human interfaces.}
  \label{fig_5}
  \end{figure}

We conduct simulation experiments on Metadrive safety benchmark\cite{metadrive}. The training session in Metadrive consists of 20 different scenarios, each featuring various typical block types and randomly placed obstacles. As shown in Fig. \ref{fig_5}, human subjects can take over control via a Logitech G29 racing wheel and monitor the training process through real-time visualization of the environment on the screen. The concepts of the observation space, action space, environmental reward, and environmental cost are as follows:\\
\indent \textbf{Observation space}: The observation space is a continuous space comprising the following elements: (a) the current state of the target vehicle, including steering angle, heading and velocity; (b) the surrounding information, represented by a 240-dimensional vector of LIDAR-like distance measurements from nearby vehicles and obstacles; (c) navigation data, including the relative positions toward future checkpoints and the destination.\\ 
\indent \textbf{Action space}: The action space is a continuous space with the acceleration and the steering angle.\\
\indent \textbf{Reward}: The reward function is composed of four parts as follows:
\begin{equation}
  R=c_{\text {disp }} R_{\text {disp }}+c_{\text {speed }} R_{\text {speed }}+c_{\text {collision }} R_{\text {collision }}+R_{\text {term }}
  \end{equation}
\indent $R_{\text {disp }}$: Encourages forward movement, defined as $R_{\text {disp }}=d_t-d_{t-1}$, where $d_t$ and $d_{t-1}$ are the longitudinal movements; $R_{\text {speed }}$: Promotes maintaining a reasonable speed, defined as $R_{\text {speed }}=v_t / v_{\max }$, where $v_t$ and $v_{\max }$ denote the current speed and maximum allowed speed; $R_{\text {collision }}$: Penalizes collisions, defined as $R_{\text {collision }}=-5$ if a collision occurs with a vehicle, human, or object, otherwise, it is 0;  $R_{\text {term }}$: If the vehicle reaches destination, $R_{\text {term }}$ is set to +10. If the vehicle drives off the road, $R_{\text {term }}$ is set to -5.\\
\indent \textbf{Cost}: Each collision with traffic vehicles or obstacles incurs a cost of -1. The environmental cost is utilized for testing the safety of the trained policies and measuring the occurrence of dangerous situations during the training process.\\
\indent We compare our approach with the following baseline methods:
\begin{itemize}
  \item Standard RL Approaches: Proximal Policy Optimization (PPO), Soft Actor-Critic (SAC), and Distributional Soft Actor-Critic (DSAC). 
  \item Offline RL Methods: Conservative Q-learning (CQL) and Imitation Learning (IL), including Behavior Cloning (BC); 
  \item HIL Approaches: Proxy Value Propagation (PVP), Human-Gated DAgger (HG-DAgger), and Intervention Weighted Regression (IWR).
\end{itemize}
\indent \indent All baseline methods are implemented using RLLib and trained on Nvidia GeForce RTX 4080 GPUs. Each experiment consists of five concurrent trials, with each trial utilizing 2 CPUs with 6 parallel rollout workers, and the experiments are repeated five times with different random seeds to ensure robustness. For H-DSAC and PVP, experiments are conducted on a local computer and repeated three times.\\
\indent The evaluation metrics are divided into two phases: training and testing. During the training phase, we focus on data usage and total safety cost, which reflects the number of collisions and potential dangers. In the testing phase, the key metrics include episodic return, episodic safety cost (average crashes per episode), and success rate (the ratio of episodes where the agent reaches the destination). For the testing phase, we use another ten different scenarios to evaluate the performance. For HIL methods, we also report human data usage and the overall intervention rate, which indicates the amount of human effort required to guide the agent.

\begin{figure}[]
  \centering
  \subfloat{\includegraphics[width=2.55in]{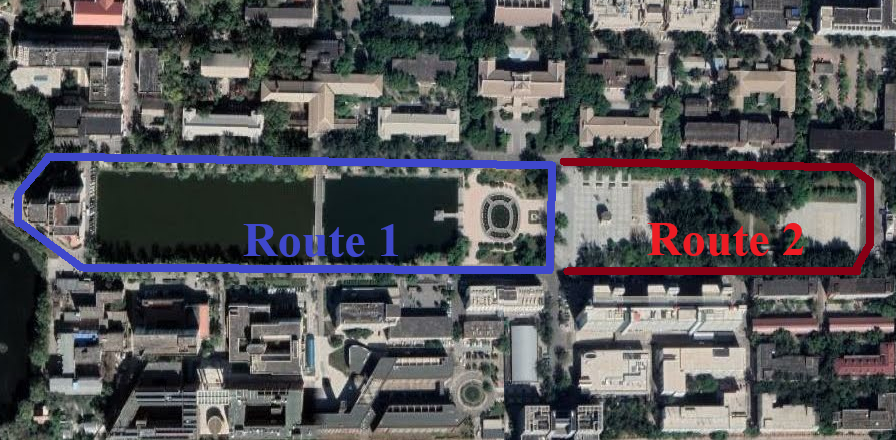}}%
  \caption{Routes for training and testing in real-world experiments.}
  \label{fig_01}
  \end{figure}
  \begin{figure}[]
    \centering
    \subfloat{\includegraphics[width=2.9in]{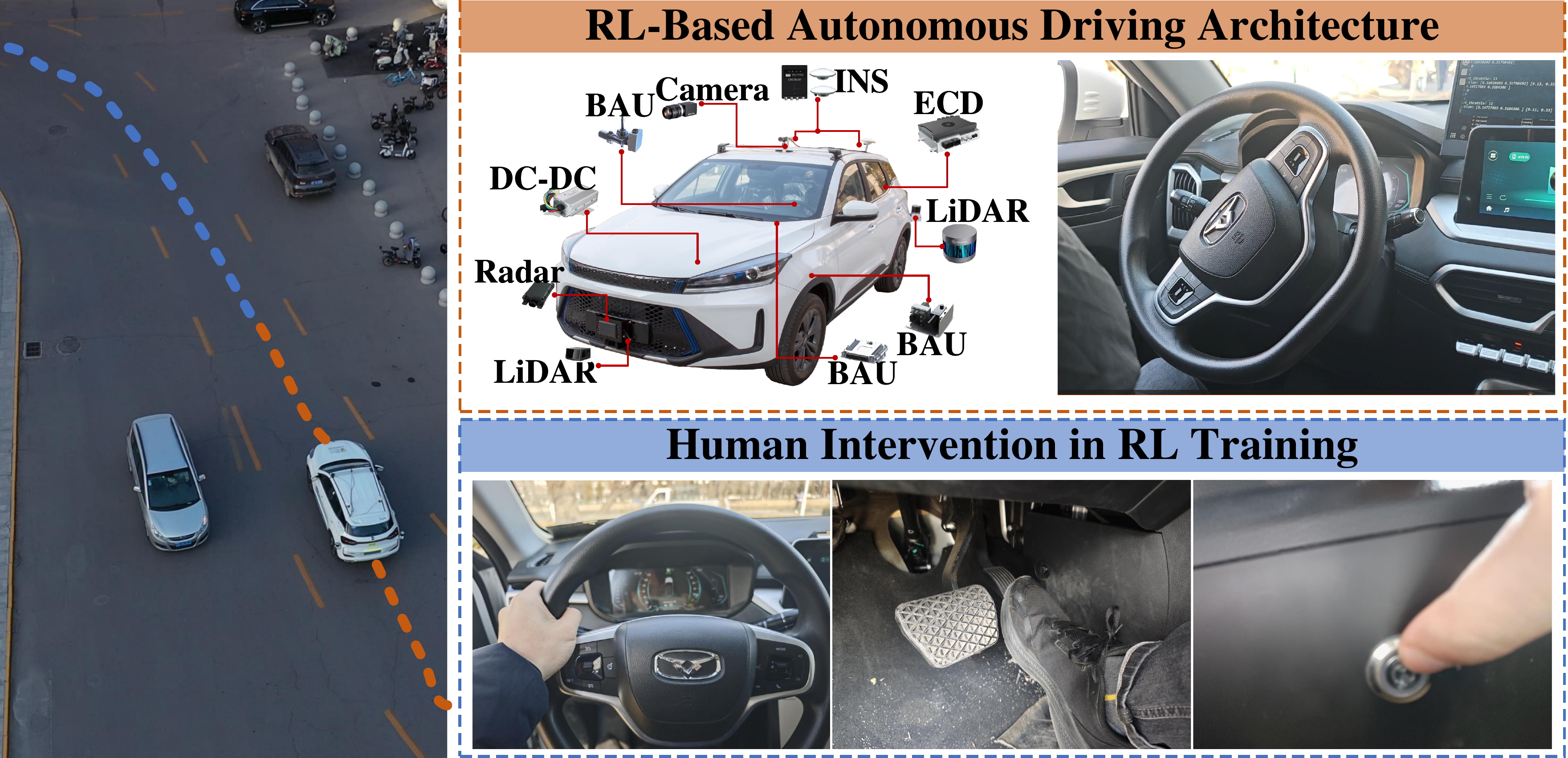}}%
    \caption{Hardware architecture and real-world setup of UGV platform.}
  \label{fig_7}
    \end{figure}

\subsection{Real-World Experiment Design}
\indent As illustrated in Fig. \ref{fig_01}, the real-world training process takes place on the campus roads of Tianjin University. Each route consists of multiple checkpoints, which specify both position and driving commands. Route 1 is used for training, while Route 2 is designated for generalization testing. The environment naturally includes random pedestrians, bicycles, and vehicles, increasing to the complexity of the training and testing conditions. The hardware architecture of our UGV platform is given in Fig. \ref{fig_7}. The localization of the UGV is achieved through an integrated navigation system (INS). For environmental perception, we utilize LiDAR, camera, and radar to detect obstacles. Object detection is performed using 3D LiDAR-based object detection and instance segmentation methods. The algorithm runs on an Nvidia Jetson AGX Orin edge computing device, while network training is conducted on a GPU. The computed control signals are then transmitted to the underlying system via a base adapter unit (BAU).\\
\indent The vehicle is trained to navigate through the checkpoints in Route 1 while actively avoiding obstacles and other vehicles. Since the H-DSAC algorithm is reward-free, there is no terminal reward, and we do not define episodes in our training process. Instead, the UGV continues driving under human supervision until a predefined number of steps is reached. Specifically, we set the total training steps to 100,000, with a policy execution frequency of 10 Hz, resulting in a total training duration of approximately two hours. Throughout the training process, human operators can intervene at any time. The driver can take control by pressing the autopilot mode switch button or using the steering wheel and throttle/brake pedals to manually override the system. This real-world experiment is designed to evaluate the practical feasibility of training an autonomous driving policy directly in real-world environments within a constrained time frame.
\begin{figure}[H]
  \centering
  \subfloat{\includegraphics[width=1.55in]{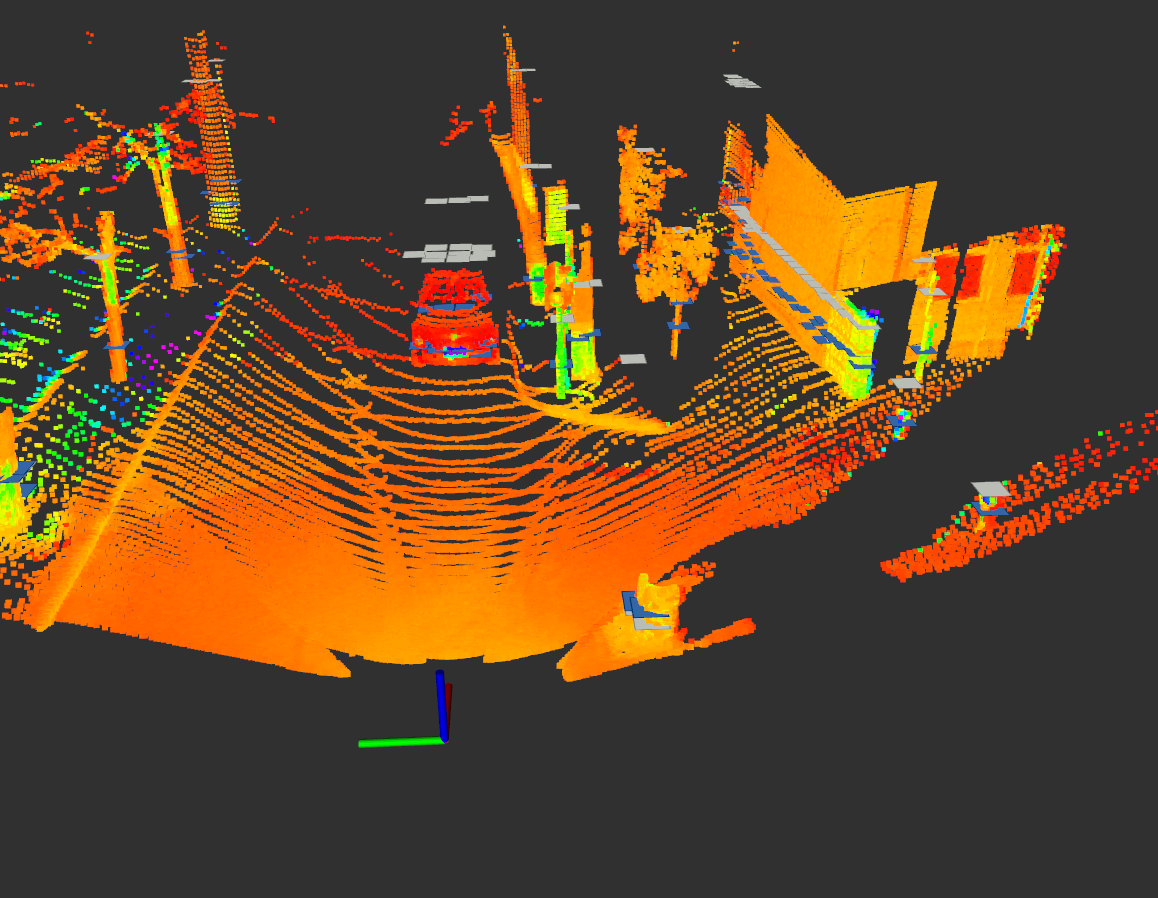}}%
  \hfil
  \subfloat{\includegraphics[width=1.2in]{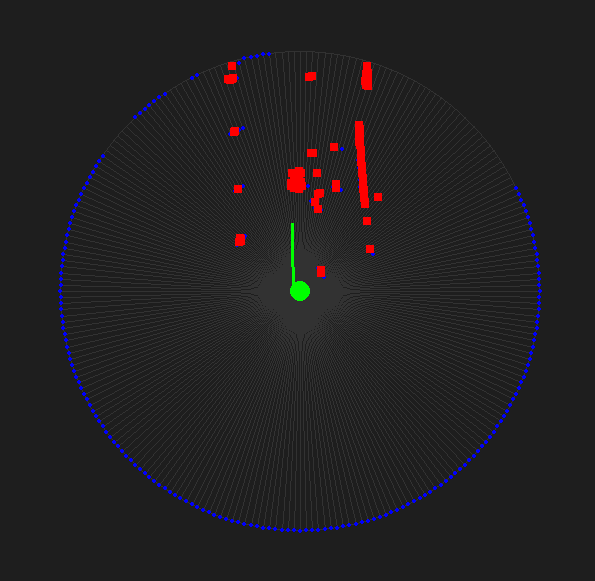}}%
  \caption{Radar-based obstacle detection and observation space visualization.}
  \label{ob}
  \end{figure}
\indent The definitions of the observation space and action space are given below:\\
\indent \textbf{Observation space}: As shown in Fig. \ref{ob}, the observation space is defined as a continuous space with the following elements: (a) Current state: Includes the target vehicle's speed, lateral offset from the lane center, and the heading angle relative to the lane center; (b) Surrounding information: We use a state representation method similar to that in MetaDrive, see Fig. \ref{ob}. Lidar is first employed to detect instances of obstacles, and this data is then converted into a vector of 240 LIDAR-like distance measurements from nearby vehicles and obstacles; (c) Navigation data: Includes the next 30 checkpoints and driving instructions, such as go straight, turn left, or turn right.\\
\indent \textbf{Action space}: The action space is defined as a continuous space with two components: acceleration and steering angle.\\
\begin{table*}[]
  \centering
  \caption{The performance of different baselines in the MetaDrive simulator.}
  \label{table1}
\begin{tabular}{lccccccc}
  \toprule
  \textbf{Method} & \multicolumn{3}{c}{\textbf{Training}} & \multicolumn{3}{c}{\textbf{Testing}} \\
  \cmidrule(lr){2-4} \cmidrule(lr){5-7}
  & \textbf{Human Data} & \textbf{Total Data} & \textbf{Safety Cost} & \textbf{Episodic Return} & \textbf{Episodic Safety Cost} & \textbf{Success Rate} \\
  \midrule
  SAC\upcite{SAC} & - & 1M & 7.94K $\pm$ 3.24K & 350.18 $\pm$ 16.21 & 1.00 $\pm$ 0.28 & 0.73 $\pm$ 0.13 \\
  PPO\upcite{PPO} & - & 1M & 45.12K $\pm$ 21.11K & 278.65 $\pm$ 35.07 & 3.92 $\pm$ 1.91 & 0.44 $\pm$ 0.14 \\
  DSAC\upcite{DSAC} & - & 1M & 7.44K $\pm$ 3.59K & 349.35 $\pm$ 22.15 & 0.47 $\pm$ 0.08 & 0.77 $\pm$ 0.09 \\
  \midrule
  Human Demo. & 50K & - & 23 & 377.523 & 0.39 & 0.97 \\
  \midrule
  CQL\upcite{CQL} & 50K (1.0) & - & - & 93.12 $\pm$ 16.31 & 1.45 $\pm$ 0.15 & 0.09 $\pm$ 0.05 \\
  BC\upcite{BC} & 50K (1.0) & - & - & 59.13 $\pm$ 8.92 & 0.12 $\pm$ 0.03 & 0 $\pm$ 0 \\
  \midrule
  HG-DAgger\upcite{DAGGEROther02} & 34.9K (0.70) & 0.05M & 56.13 & 142.35  & 2.1 & 0.30 \\
  IWR\upcite{IWR} & 37.1K (0.74) & 0.05M & 48.78 & 329.97  & 4.00 & 0.70 \\
  \midrule
  PVP\upcite{PVP} & 15.7K & 0.05M & 33.67 $\pm$ 3.46 & 338.28 $\pm$ 10.21 & 0.65 $\pm$ 0.12 & 0.80 $\pm$ 0.03 \\
  H-DSAC (Ours) & 14.8K  & 0.05M & \textbf{32.12 $\pm$ 4.68} & \textbf{353.39 $\pm$ 12.34} & \textbf{0.31 $\pm$ 0.03} & \textbf{0.83 $\pm$ 0.05} \\
  \bottomrule
  \end{tabular}
\end{table*}
\section{Experiment Results}
\subsection{Simulation Experiment Results}
\begin{figure}[]
  \centering
  \subfloat{\includegraphics[width=1.6in]{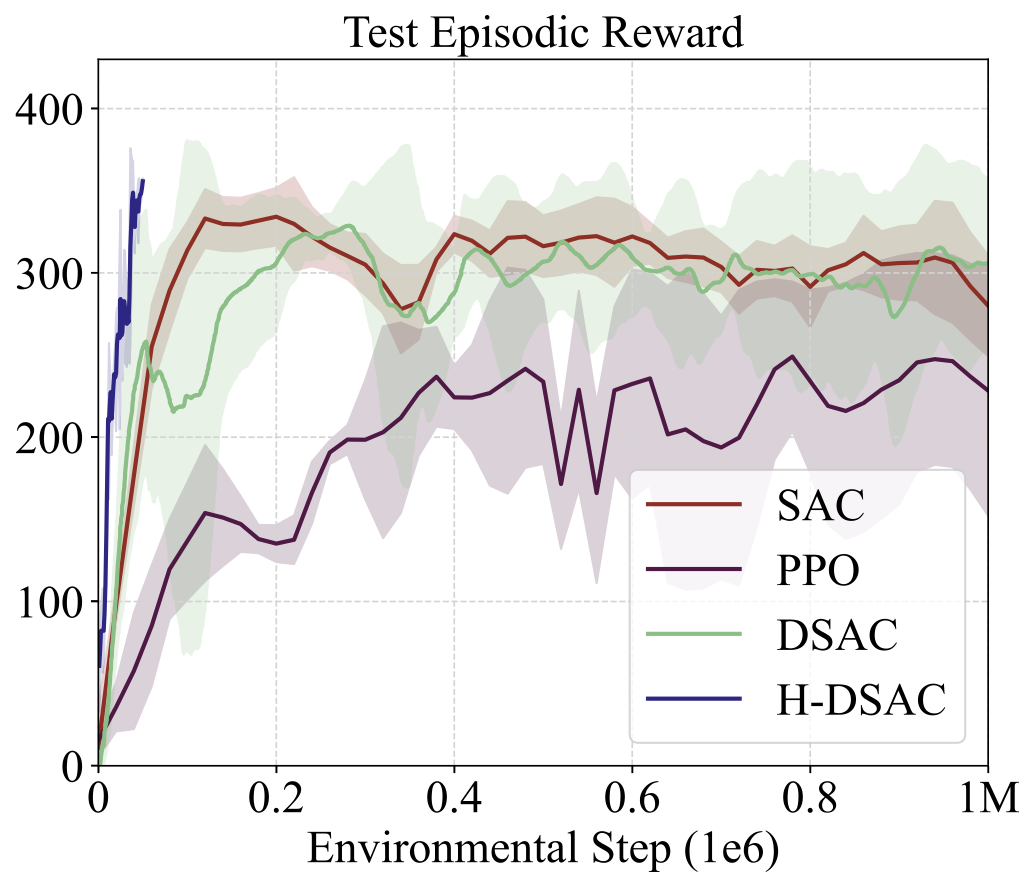}}%
  \hfil
  \subfloat{\includegraphics[width=1.6in]{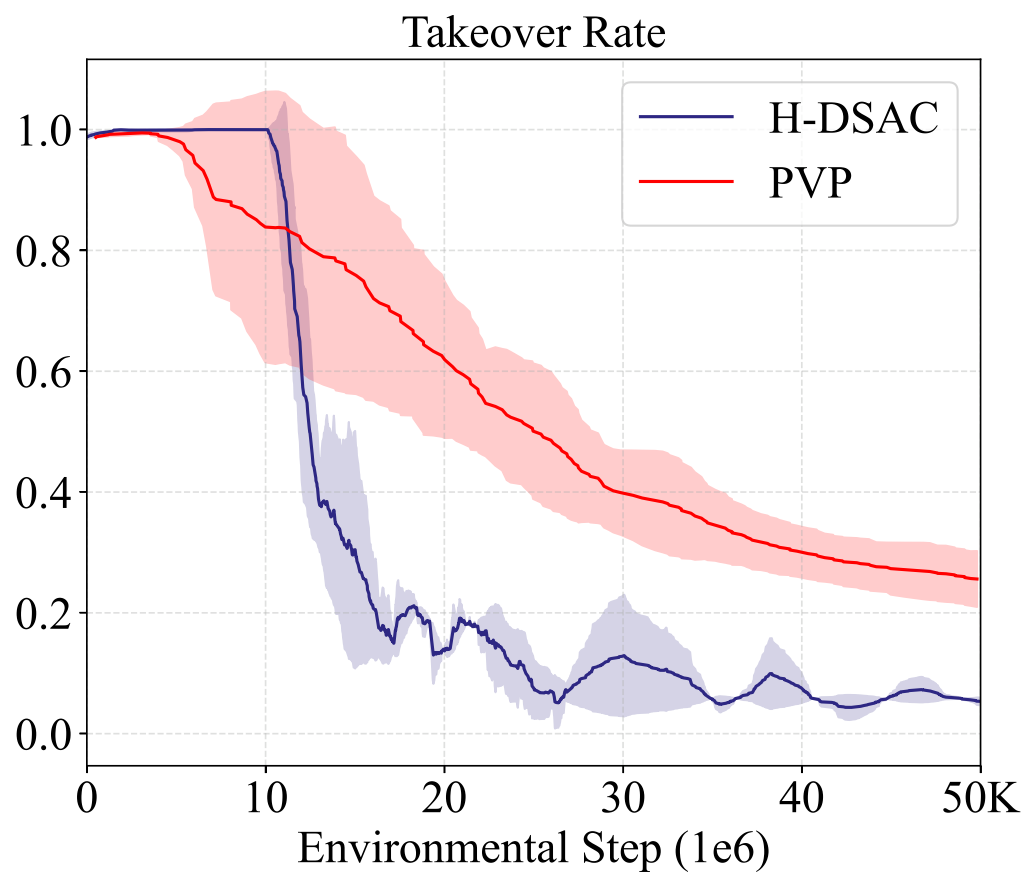}}%
  \caption{Comparison of different baselines in the simulation experiment.}
  \label{test_r}
  \end{figure}

The performance of different baselines is summarized in Table \ref{table1}, with learning curve illustrated in Fig. \ref{test_r}.

\indent As shown in Table \ref{table1}, H-DSAC outperforms standard RL algorithms such as SAC, PPO, and DSAC. It achieves a higher episodic return (353.39) and a lower episodic safety cost (0.31) compared to SAC (350.18, 1.00), PPO (278.65, 3.92), and DSAC (349.35, 0.47). It also maintains the highest success rate (0.83) among the RL methods. When compared to offline RL (CQL) and IL methods (BC), H-DSAC demonstrates superior performance, significantly outperforming CQL (93.12 return, 9\% success rate) and BC (59.13 return, 0\% success rate). This highlights its ability to generalize effectively while ensuring safety.
\begin{figure}[]
  \centering
  \subfloat{\includegraphics[width=1.6in]{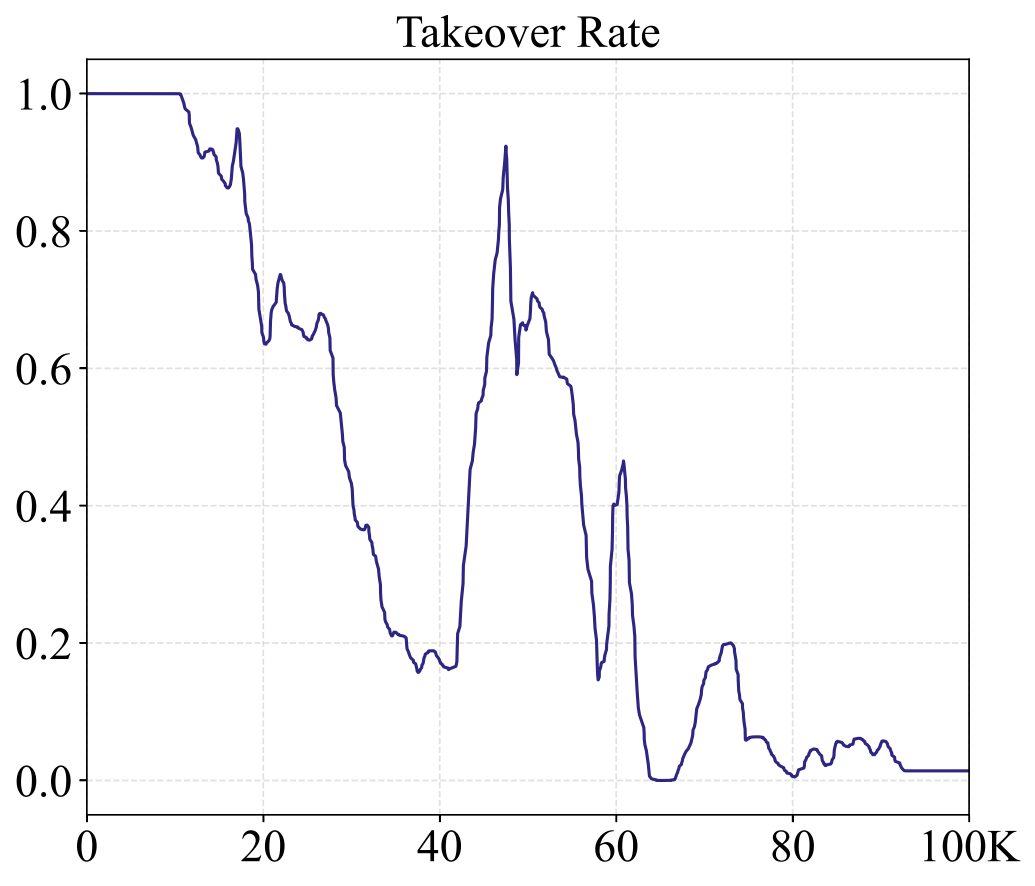}}%
  \hfil
  \subfloat{\includegraphics[width=1.6in]{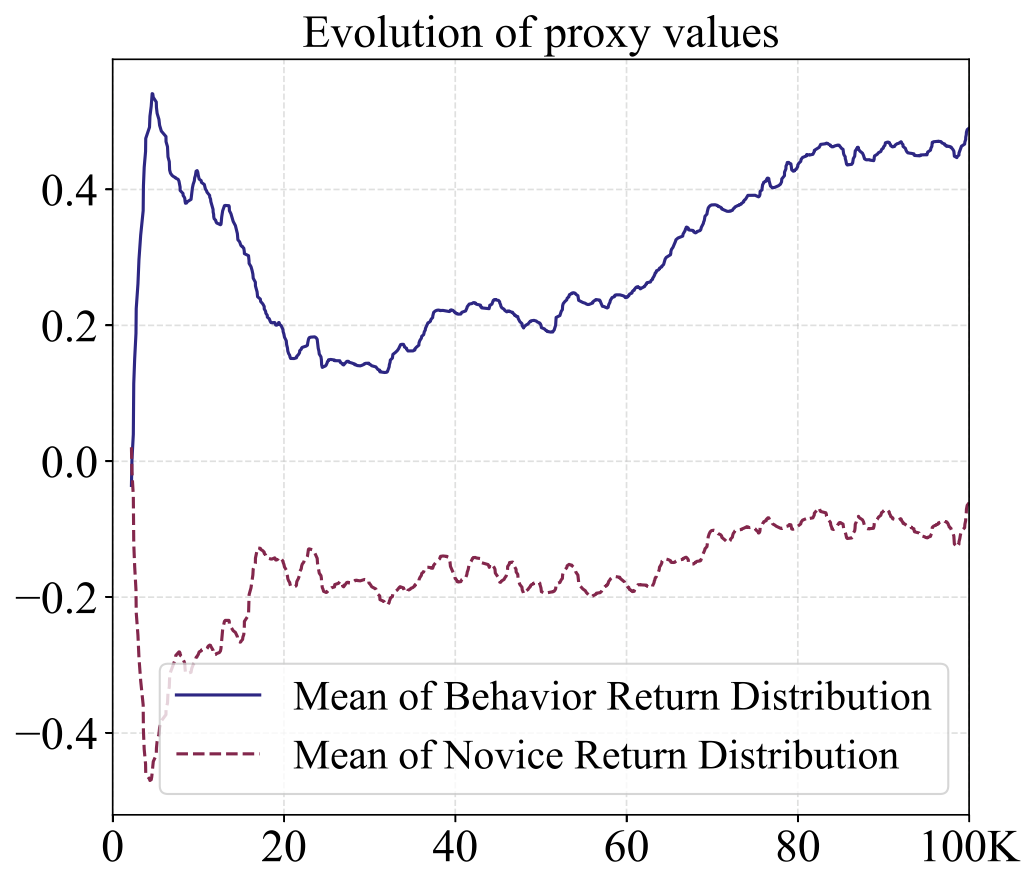}}%
  \caption{Takeover rate and proxy values in the real-world training.}
  \label{real_train}
  \end{figure}

  \begin{figure*}[!t]
    \centering
    \subfloat{\includegraphics[width=1.73in]{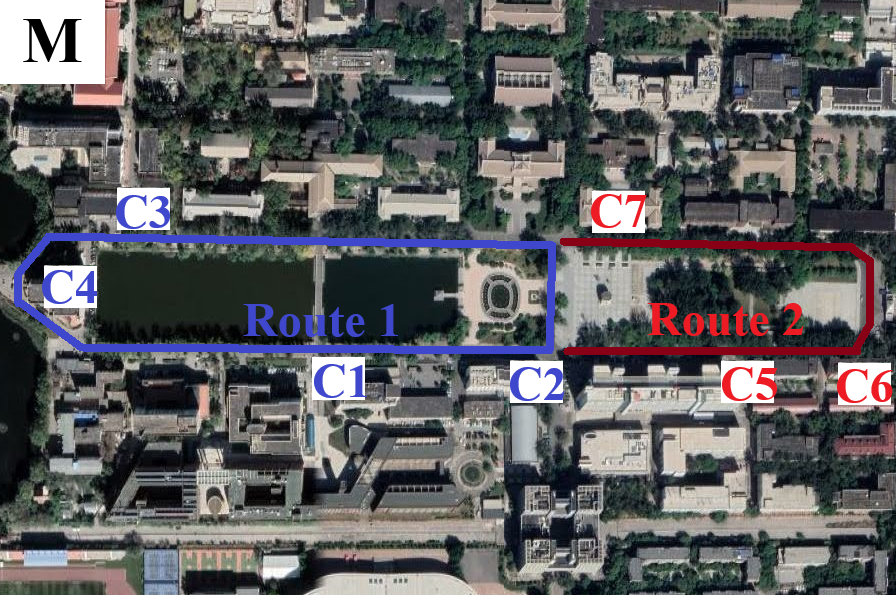}}
    \hfil
    \subfloat{\includegraphics[width=1.73in]{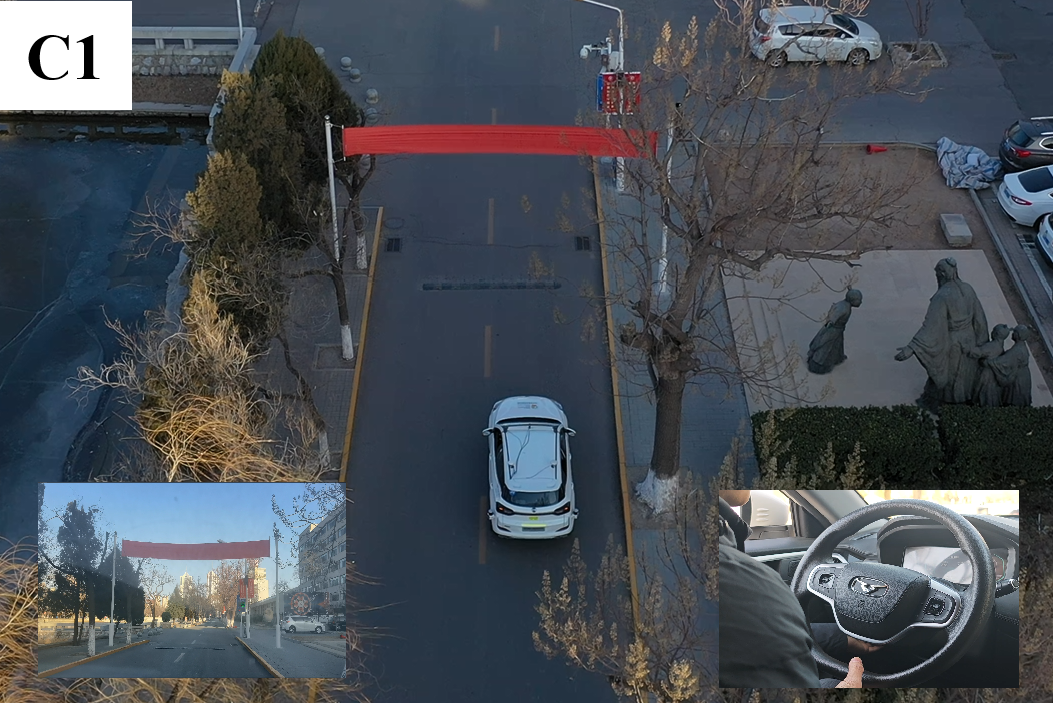}}
    \hfil
    \subfloat{\includegraphics[width=1.73in]{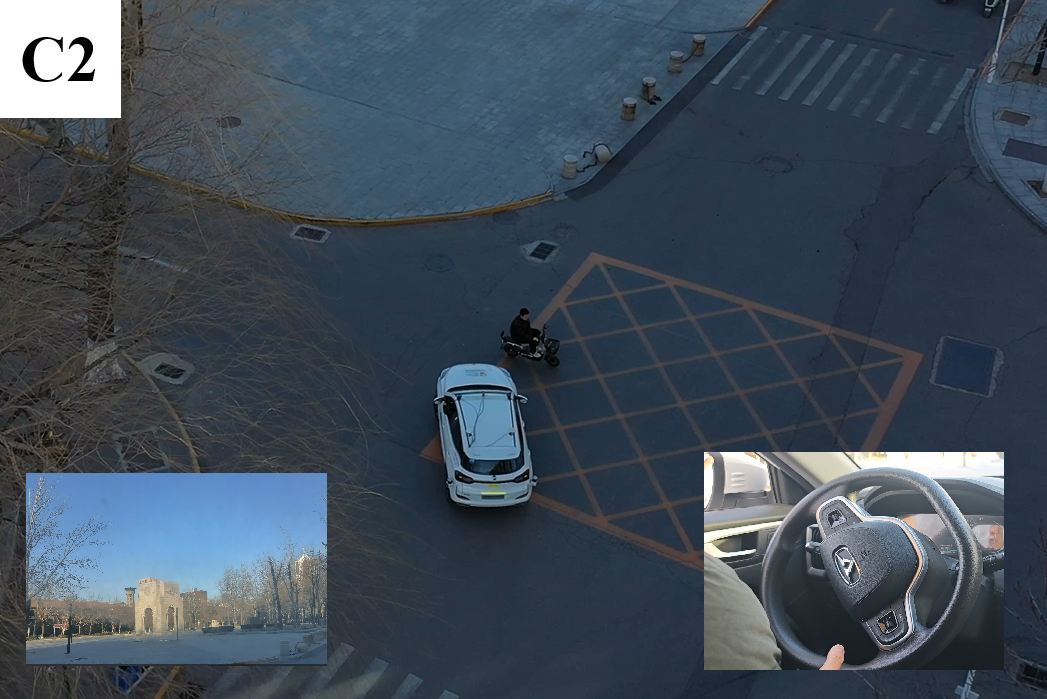}}
    \hfil
    \subfloat{\includegraphics[width=1.73in]{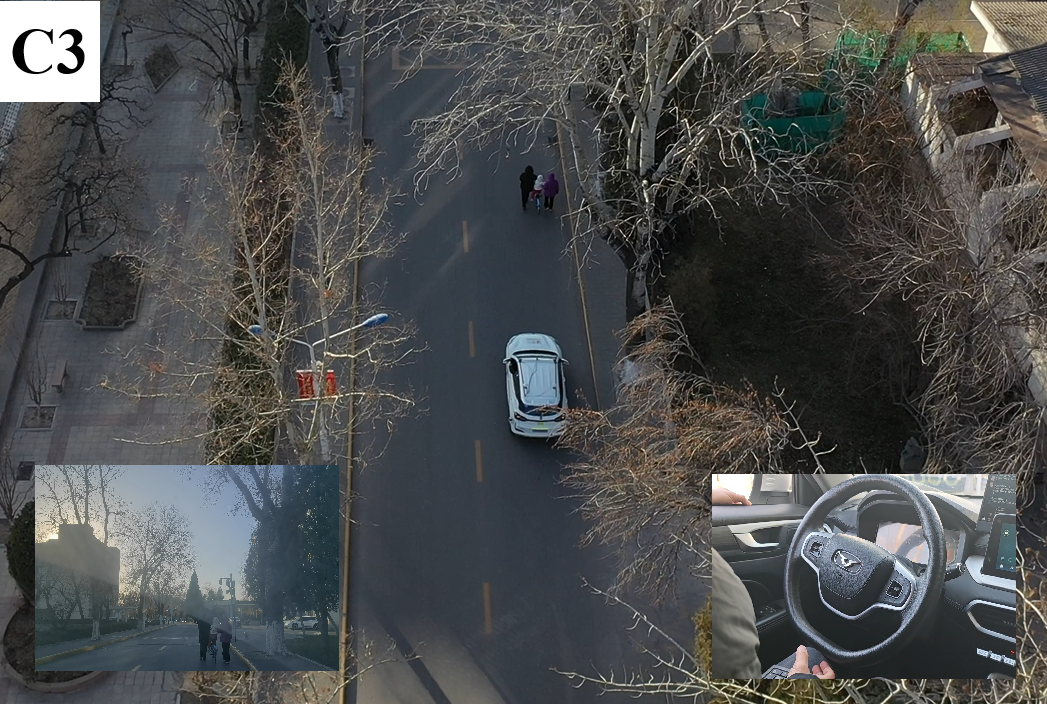}}
    \\
    \subfloat{\includegraphics[width=1.73in]{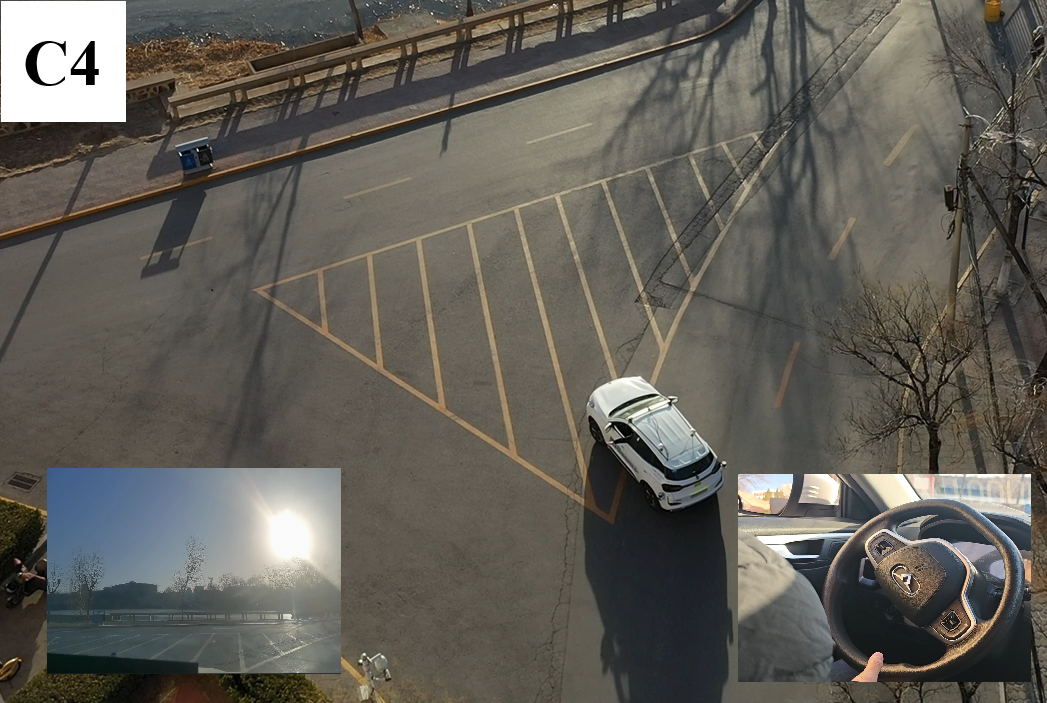}}
    \hfil
    \subfloat{\includegraphics[width=1.73in]{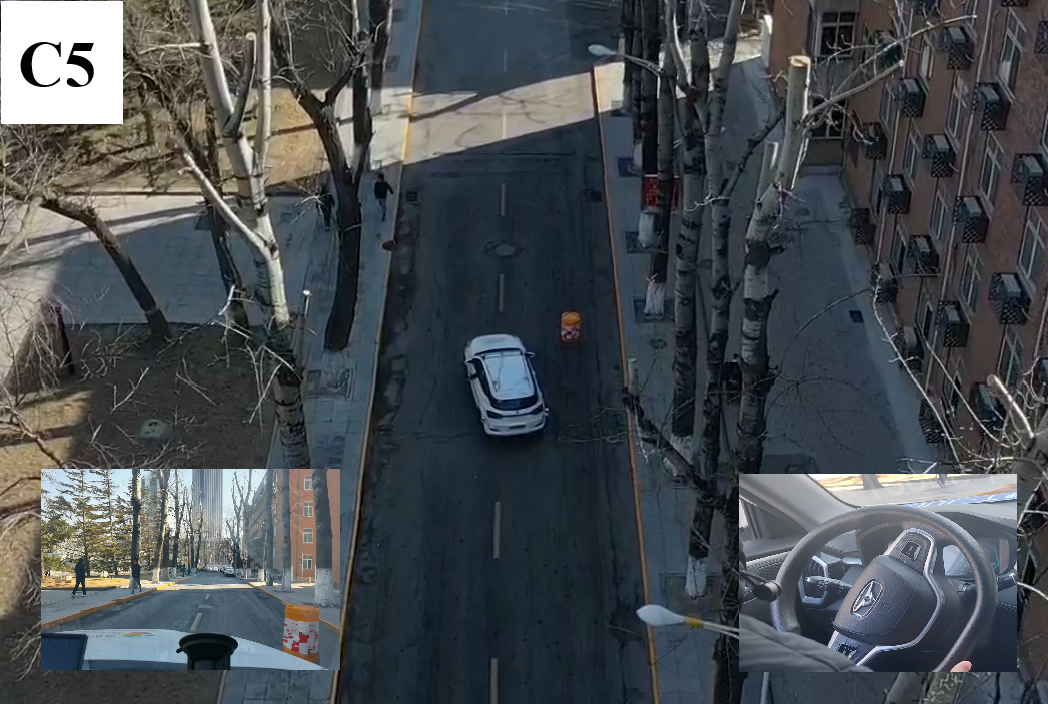}}
    \hfil
    \subfloat{\includegraphics[width=1.73in]{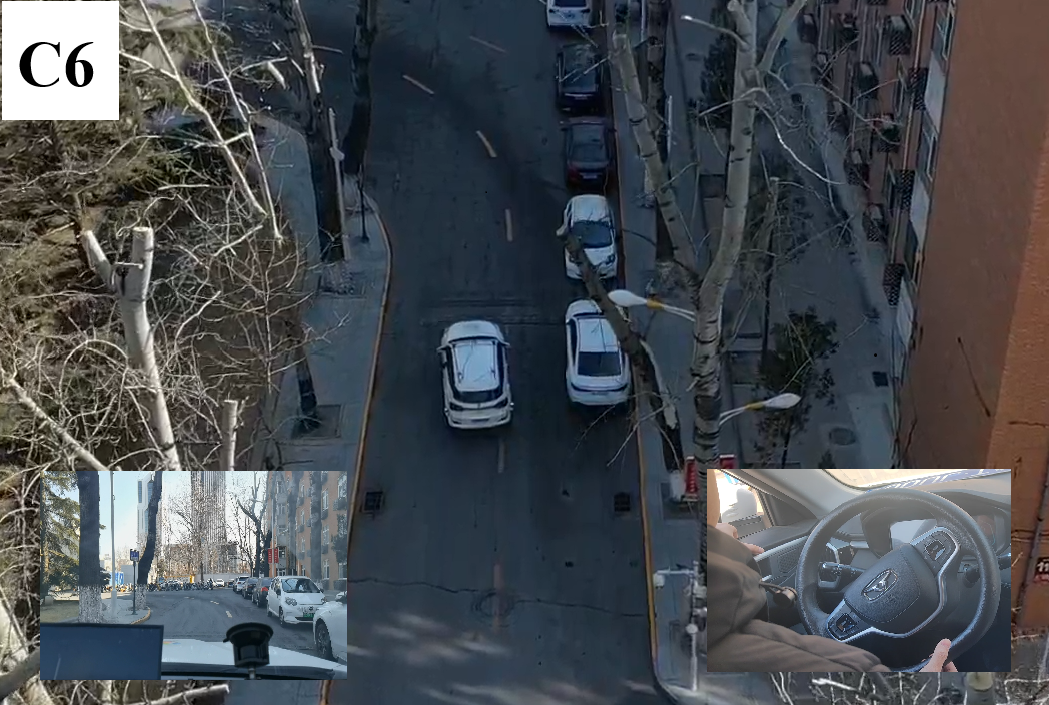}}
    \hfil
    \subfloat{\includegraphics[width=1.73in]{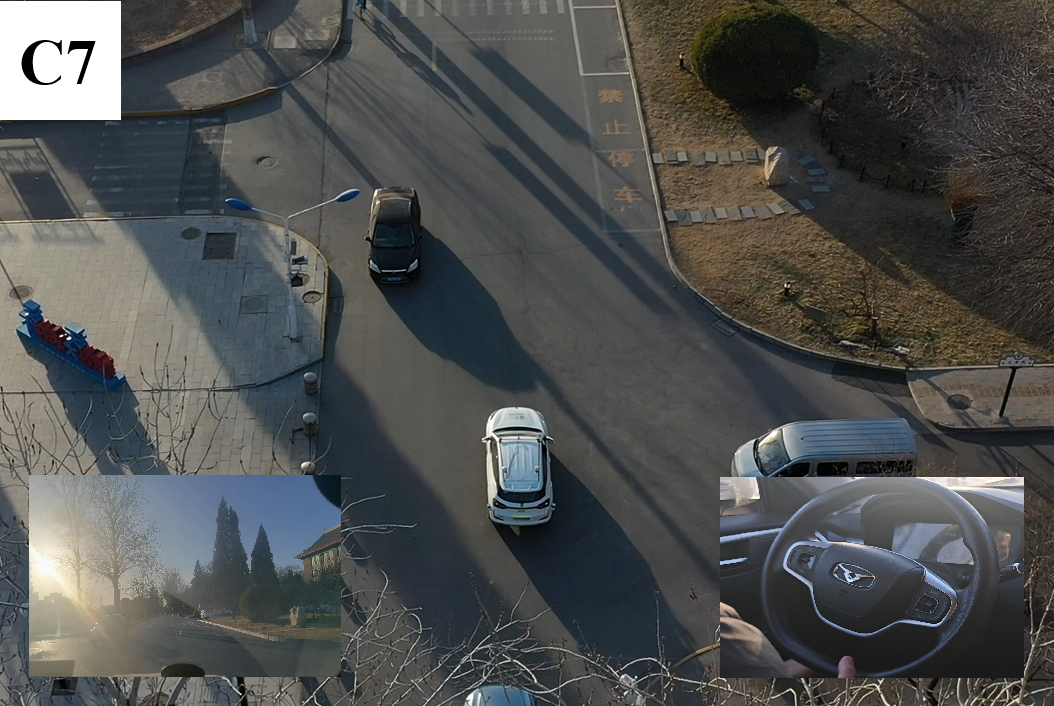}}
    \caption{Details fo real-world experiment. (M) Routes for training and testing. (C1-C4) Real-world driving performance on Route 1. (C5-C7) Real-world driving performance on Route 2.}
    \label{final}
    \end{figure*}
\indent Among other HIL methods like HG-Dagger and IWR, H-DSAC achieves the highest success rate (83\%) and the lowest safety cost (32.12). Compared to PVP, H-DSAC exhibits faster convergence and higher performance, with a final success rate of 83\% versus PVP's 80\%, while maintaining a comparable amount of human data. These results underscores H-DSAC’s efficiency in leveraging human guidance to enhance both safety and performance.
\subsection{Real-World Experiment Results}
As illustrated in Fig. \ref{real_train}, during the initial phase (0 to 10k steps), the vehicle's behavior is highly random due to the untrained policy, leading to poor performance and frequent human takeovers. From 10k to 40k steps, the system gradually improves, with the distributional proxy value function loss decreasing and the takeover rate dropping. The vehicle begins to drive straight but remains unstable with noticeable speed oscillations. At around 50k steps, the introduction of more complex scenarios (e.g., pedestrians, cyclists, and surrounding vehicles) causes a temporary spike in the takeover rate, as the vehicle struggles to handle these challenges. By 60k steps, the system adapts, and the takeover rate decreases again, indicating improved robustness. By 80k steps, the policy stabilizes, and the vehicle is able to independently complete the route without human intervention.\\
\begin{figure}[]
  \centering
  \subfloat{\includegraphics[width=1.7in]{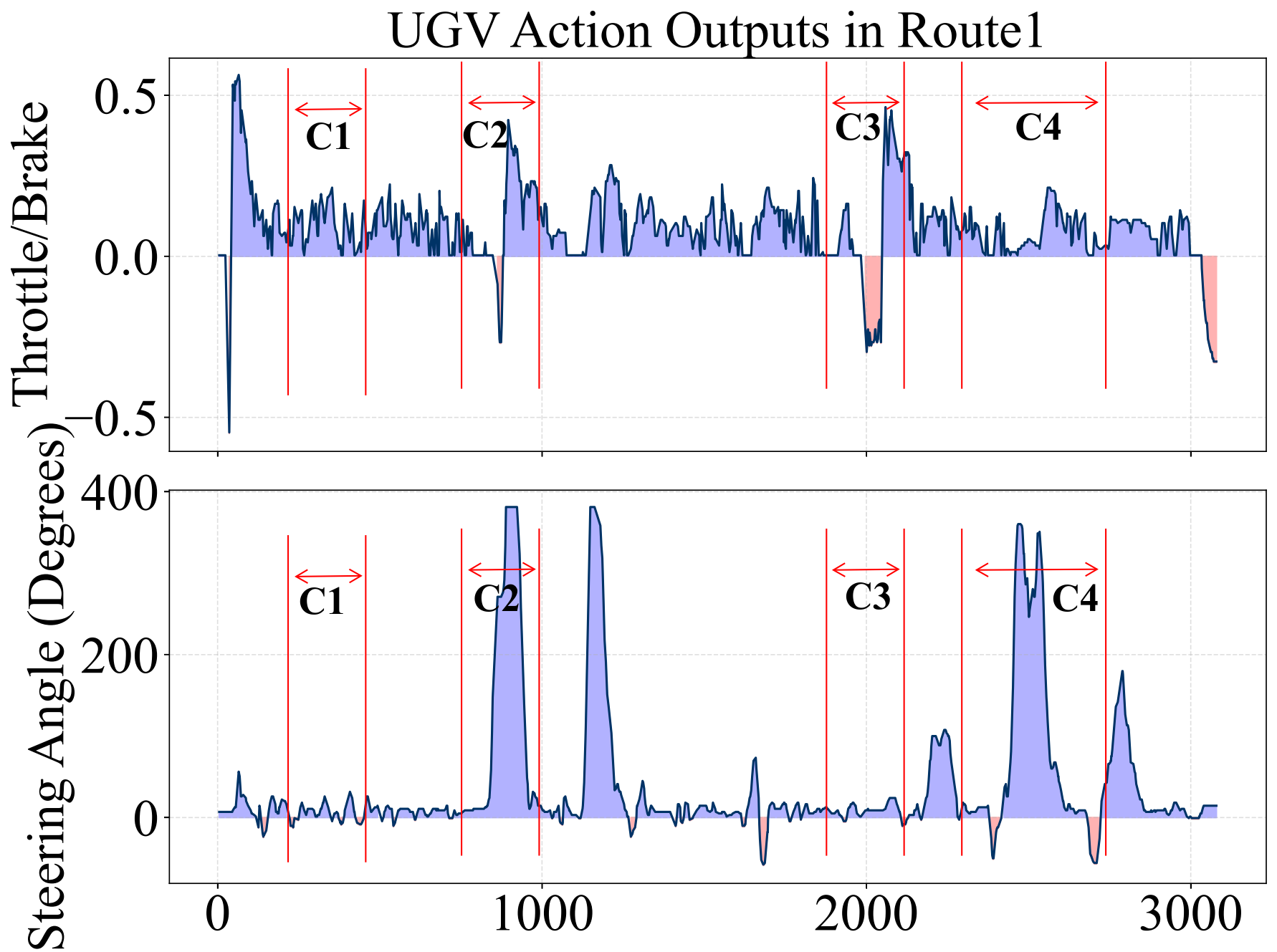}}%
  \hfil
  \subfloat{\includegraphics[width=1.7in]{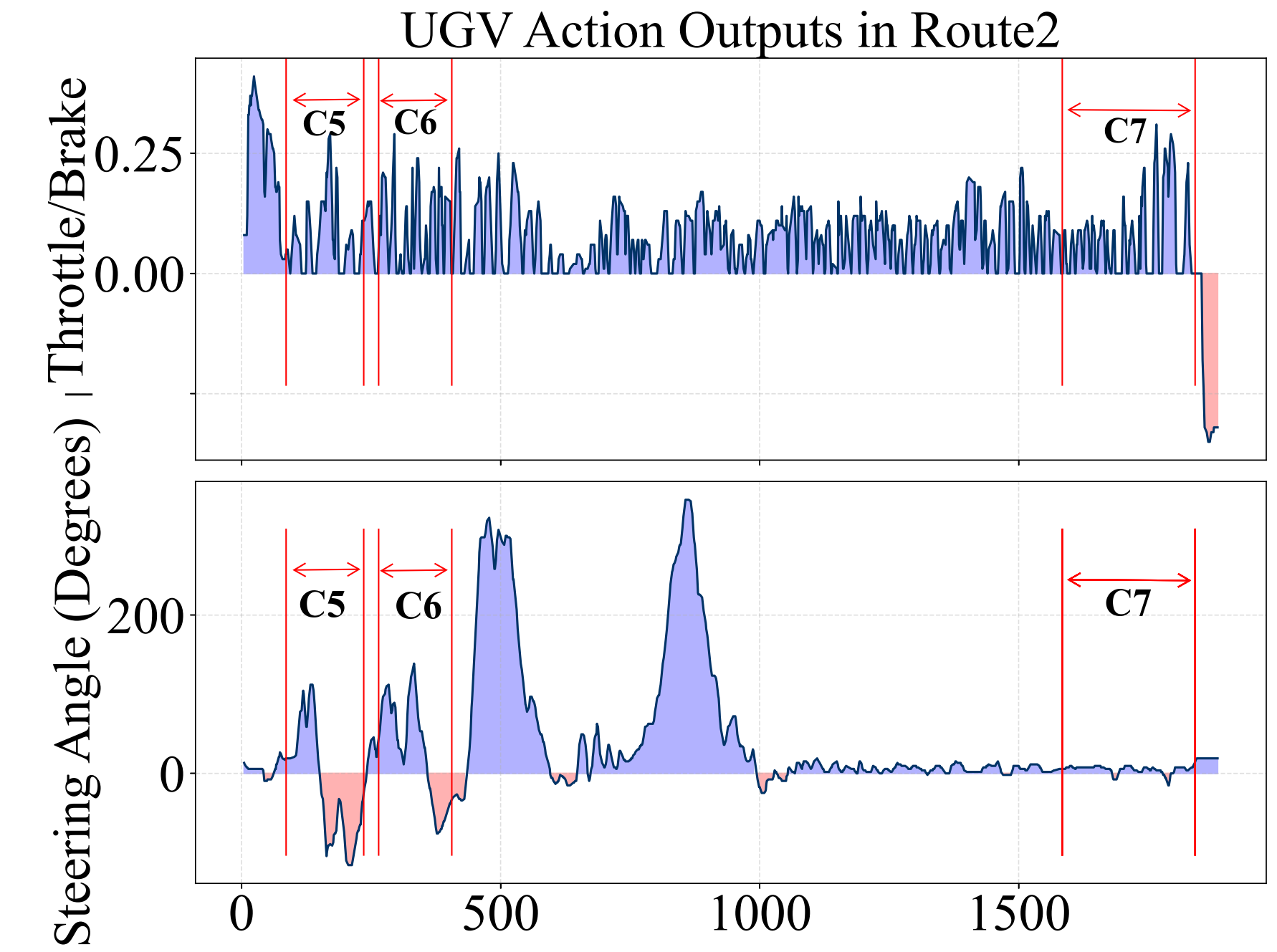}}%
  \caption{Action outputs of the vehicle in real-world scenarios.}
  \label{action}
  \end{figure}
\indent As shown in Fig. \ref{final}, the vehicle is trained on Route 1 and subsequently tested on both Route 1 and Route 2. During testing on Route 1, the vehicle successfully completes the entire route. As illustrated in Fig. \ref{final}(C1), it maintains a stable lane position while driving straight. In Fig. \ref{final}(C2), it executes a left turn while avoiding a pedestrian, and in Fig. \ref{final}(C3), it slows down and stops to yield to a crossing pedestrian. Additionally, as shown in Fig. \ref{final}(C4), the vehicle successfully executes a sharp turn. The corresponding action outputs are presented in Fig. \ref{action}. To evaluate generalization, the vehicle is also tested on Route 2, with its action outputs given in Fig. \ref{action}. On this route, as depicted in Fig. \ref{final}(C5), it successfully maneuvers around an obstacle. In Fig. \ref{final}(C6), it navigates past a stationary vehicle, and in Fig. \ref{final}(C7), it effectively manages an intersection with heavy traffic.\\
\indent These results demonstrate that H-DSAC can learn driving policies in real-world environments with high sample efficiency and low safety costs. The vehicle handles complex scenarios and exhibits strong generalization capability.

\section{CONCLUSION}
This paper presented the human-guided distributional soft actor-critic (H-DSAC), a novel reinforcement learning approach that integrates human feedback to enhance sample efficiency, safety, and performance in real-world autonomous driving. By leveraging human guidance through proxy value propagation, H-DSAC efficiently trained the agent to navigate complex environment with minimal need for explicit reward engineering. This ensures safe and robust learning. Experimental results from both simulation and real-world environments demonstrated that H-DSAC outperformed standard RL, offline RL, imitation learning, and other HIL methods in terms of return, safety, and success rate. These findings highlighted the potential of H-DSAC to enable efficient real-world autonomous driving policy learning within practical training times, showcasing its ability to balance human expertise with autonomous exploration for safe and effective driving.

\bibliographystyle{IEEEtran}
\bibliography{references}

\end{document}